\documentclass[runningheads]{llncs}

\usepackage{eccv}

\usepackage{eccvabbrv}

\usepackage[dvipsnames]{xcolor}
\usepackage{multirow}
\usepackage{multicol}
\usepackage{colortbl}
\usepackage{amsmath}
\usepackage{amssymb}
\usepackage{algorithm}
\usepackage{setspace}
\usepackage{algpseudocode}
\usepackage{listings}
\definecolor{codegreen}{rgb}{0,0.5,0}
\definecolor{codeblue}{rgb}{0.25,0.5,0.5}
\definecolor{codegray}{rgb}{0.6,0.6,0.6}
\usepackage{pifont}% http://ctan.org/pkg/pifont
\newcommand{\cmark}{\ding{51}}%
\usepackage{graphicx}
\usepackage{amsmath}
\usepackage{amssymb}
\usepackage{booktabs}
\usepackage{array}
\usepackage{multirow}
\usepackage{makecell}
\usepackage[dvipsnames]{xcolor}

\usepackage{float}
\usepackage{marvosym}

\usepackage[accsupp]{axessibility}  % Improves PDF readability for those with disabilities.

\usepackage{hyperref}

\usepackage{orcidlink}

\begin{document}

\title{PPAD: Iterative Interactions of Prediction and Planning for End-to-end Autonomous Driving} 

\titlerunning{PPAD}

\author{
Zhili Chen\inst{1\dagger}\orcidlink{0000-0002-8272-156X} \and
Maosheng Ye\inst{1}\orcidlink{0000-0001-8470-685X} \and
Shuangjie Xu\inst{1}\orcidlink{0000-0003-0150-7068} \and \\
Tongyi Cao\inst{2}\orcidlink{0000-0003-2157-1526} \and 
Qifeng Chen\inst{1}\textsuperscript{\Letter}\orcidlink{0000-0003-2199-3948}
}

\authorrunning{Z.~Chen et al.}

\institute{
$^1$HKUST \qquad
$^2$DeepRoute.AI \\
\email{\{zchenei, myeag, shuangjie.xu\}@connect.ust.hk}, \\
\email{tongyicao@deeproute.ai}, \email{cqf@cse.ust.hk}
}

\renewcommand{\thefootnote}{}
\footnotetext{$^\dagger$Work done during an internship at DeepRoute.AI.\\
\textsuperscript{\Letter}Corresponding author.}

\maketitle
\begin{abstract}
We present a new interaction mechanism of prediction and planning for end-to-end autonomous driving, called PPAD (Iterative Interaction of \textbf{P}rediction and \textbf{P}lanning \textbf{A}utonomous \textbf{D}riving), which considers the timestep-wise interaction to better integrate prediction and planning. An ego vehicle performs motion planning at each timestep based on the trajectory prediction of surrounding agents (e.g., vehicles and pedestrians) and its local road conditions. Unlike existing end-to-end autonomous driving frameworks, PPAD models the interactions among ego, agents, and the dynamic environment in an autoregressive manner by interleaving the \textbf{Prediction} and \textbf{Planning} processes at every timestep, instead of a single sequential process of prediction followed by planning. Specifically, we design ego-to-agent, ego-to-map, and ego-to-BEV interaction mechanisms with hierarchical dynamic key objects attention to better model the interactions. The experiments on the nuScenes benchmark show that our approach outperforms state-of-the-art methods. Project page at \url{https://github.com/zlichen/PPAD}.
\end{abstract}
\section{Introduction}
\label{sec:intro}

The blossom of deep learning techniques has empowered autonomous driving, where many exciting milestones in autonomous driving have burst into our eyes owing to the convenient and interpretable discrete module designs. Recently, the planning-oriented~\cite{hu2023planning} philosophy resonated with the community for pursuing a more effective end-to-end driving system, which is the focus of this work.

\begin{figure}[t!]
\centering
\includegraphics[width=1.0\linewidth]{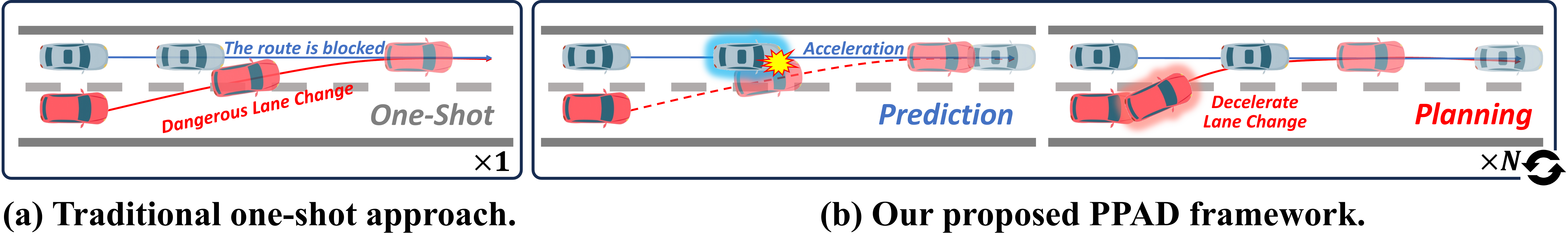}

        \caption{
        \textbf{A high-level illustration of our proposed PPAD framework.} The agent (in \textcolor{blue}{blue}) intends to drive straight, while the ego (in \textcolor{red}{red}) plans to change lanes. Fig.~\ref{fig:coreIdea}(a) presents the typical one-shot method that might result in invalid motion plans and lead to an accident because of a lack of in-depth interactions. Fig.~\ref{fig:coreIdea}(b) demonstrates the gaming process between the ego and the agent under the PPAD architecture. During the prediction process, the agent executes an assertive plan by accelerating to stop the ego from blocking its route. The planning process of the ego plans trajectory based on the previous prediction process of the agent. The ego decelerates to avoid a potential accident and then changes lanes to achieve its driving goal.
        }
        \label{fig:coreIdea}
\end{figure}

Traditional methods in an autonomous driving system often break down the system into modular components, including localization, perception, tracking, prediction, planning, and control for interpretability and visibility. However, there are several drawbacks: 1) the accumulation of errors between modules becomes more significant as the system complexity increases. 2) the performance of the downstream task is highly related to the upper stream module, which makes it very difficult to construct a unified data-driven infrastructure. 

Recently, end-to-end autonomous driving has gained popularity due to its simplicity. Two main lines are proposed based on the learning architecture. The first kind of method~\cite {codevilla2019exploring} takes the raw sensor data as input and directly outputs the planning trajectories or control command without any view transformation as intermediate representations for scene understanding. The other kinds of approaches~\cite{hu2023planning,jiang2023vad} are built upon BEV representation and fully utilize the queries to generate the intermediate outputs as guidance for producing the planning results. One of the most significant advantages lies in the interpretability. In this work, we follow the design of the second kind of work. 

VAD~\cite{jiang2023vad} and UniAD~\cite{hu2023planning} are typical one-shot motion planning methods, which only consider a single-step interaction between agents, ego-agent and surrounding environment (e.g., map elements). ThinkTwice~\cite{jia2023think} makes it a two-stage framework to enhance the gaming or interaction procedure. QCNet~\cite{zhou2023query} and GameFormer~\cite{huang2023gameformer} also recurrently model the trajectory prediction task. Given motion planning is a computational problem that finds a sequence of valid trajectories, often based on surrounding agents' forecasting, environmental understanding, and historical and future contexts. It can also be viewed as a game in which agents continuously plan their next move according to other agents' intentions and the encountering environment, further achieving their ultimate goals through incremental actions. To model these dynamic interactions of prediction and planning in end-to-end autonomous driving, it is crucial to consider the possible variance of predicted trajectories through multi-step modeling for planning feasible trajectories.

Inspired by VAD~\cite{jiang2023vad}, we aim to introduce the step-by-step Prediction-Planning into a learning-based framework. Intuitively, the prediction and planning modules can be modeled as a motion forecasting task, which predicts future waypoints by the given historical information. The results of prediction and planning modules at each time step are highly dependent on each other. Therefore, we need to consider the agent-agent and agent-environment interactions iteratively and bidirectionally to maximize the expectation of agents' prediction under the given observation of the other agents. We propose our \textbf{PPAD} to plan the ego agent's future trajectories step-by-step to model the timestep-wise bidirectional interaction or gaming in a vectorized learning framework as shown in Fig.~\ref{fig:coreIdea}. PPAD consists of the prediction and planning process. For each motion forecasting step, 1) \textbf{Prediction} process generates current step motion states by cross-attention and self-attention among agents and environment based on previous motion states to model the fine-grained bidirectional interactions. We take ego-agent-environment-BEV interaction into account to propagate features among all the traffic participants. 2) \textbf{Planning} process predicts the current step motion trajectories based on the expectation process. Our contributions are summarized as follows:

\begin{itemize}
 \renewcommand{\labelitemi}{\textbullet}
    \item We propose PPAD that optimizes ego-agent-environment interactions in an iterative prediction-planning manner. Iterative optimization could model the interactions and gaming better and more naturally in a planning task. The prediction process deals with more fine-grained and complex future uncertainties for multi-agent context learning, while the planning process plans a one-step future trajectory for the ego vehicle. 
    \item We model fine-grained interactions among the ego vehicle, agents, environment, and BEV features map, step-by-step with hierarchical dynamic key objects attention emphasizing on the spatial locality.
    \item The experiments conducted on the nuScenes~\cite{caesar2020nuscenes} and Argoverse~\cite{chang2019argoverse,wilson2023argoverse} datasets have demonstrated the effectiveness of our approach over state-of-the-art approaches.
\end{itemize}

\section{Related Work}
\subsection{Multi-stage Autonomous Driving} Most autonomous driving systems are built upon the multi-stage design philosophy, which commonly consists of localization, perception, and planning. The perception module has been well studied recently due to the emergence of deep learning. Camera-based~\cite{li2022bevformer,huang2021bevdet,li2023bevdepth}, Lidar-based~\cite{ye2020hvnet,ye2021tpcn,ye2021drinet,langPointPillarsFastEncoders2019,zhouVoxelNetEndtoEndLearning2018} or fused-based~\cite{liu2023bevfusion,xie2023sparsefusion} approaches are proposed to fully exploit the potential of raw sensor data in order to produce accurate 3D objects prediction, semantic segmentation or tracking velocity. Prediction takes the outputs of the perception module to generate the future waypoints for the ego and the agents. Current approaches~\cite{bansal2018chauffeurnet,chai2019multipath,duvenaud2015convolutional,gao2020vectornet,henaff2015deep,liang2020learning,liu2021multimodal,shuman2013emerging,song2022learning,ye2021tpcn,zeng2021lanercnn,zhou2022hivt,fan2018baidu,zhou2023query} explore different representations to encode surrounding environment (map information) and agent interactions to predict final trajectories by regression or postprocessing sampling strategies. Some other works~\cite{luo2018fast,casas2018intentnet} propose a joint perception and prediction framework, which aggregates historical information to generate tracklets with future trajectories. This unified learning framework could help address the non-differential process and alleviate the unstable perception problem, compared with previous works. Based on the perception and prediction results, planning module~\cite{fan2018baidu,chekroun2021gri,renz2022plant,bansal2018chauffeurnet} plan its future behavior by cost-map optimization or learning-based approaches.

\subsection{End-To-End Autonomous Driving} Recently, more and more works have focused on end-to-end autonomous driving due to its merits in reducing internal accumulative errors and direct yet simple learning objectives. Typical methods~\cite{wu2022trajectory,bojarski2016end} take the raw visual inputs to regress the final control command or trajectory points without any view transformations. To embrace reinforcement learning, a series of following works~\cite{zhang2021end,wu2022trajectory,chen2020learning} utilize policy-based or valued-based approaches to improve driving behavior. With the popularity of BEV representation, more advanced architectures~\cite{jia2023think,chitta2022transfuser} are introduced to attach more interpretability and help deal with the complex interactions in the driving scenarios. Another merit of BEV representation lies in its simplicity in fusing multi-modality sensors. Moreover, the modularized approaches~\cite{hu2023planning,jiang2023vad,hu2022st,gu2023vip3d} decouple the end-to-end learning-based methods into several submodules or subtasks, while in a multi-task learning manner. The unified design could propagate and share learning context between modules through queries or feature maps.

\subsection{BEV Representation} BEV representation has gained significant prominence in the field of autonomous driving systems due to its inherent distortion-free characteristics and its simplicity in facilitating multi-sensor fusion. There are two main lines for BEV representation, including bottom-up and top-down ways. LSS~\cite{philion2020lift} stands out as a bottom-up pioneering work that explores depth distribution for the 3D space frustum sampling to form BEV representation. Works~\cite{li2023bevdepth,huang2021bevdet} optimize the pipeline through better depth estimation or lightweight sampling design. BEVFormer~\cite{li2022bevformer} and its following works~\cite{liu2022petr,liu2022petrv2,huang2021bevdet,wang2023exploring,yang2023bevformer,liu2023vision} adopt the top-down architecture, which uses a deformable transformer for the view transformations without depth supervision. 

\begin{figure}[t!]
\centering
\includegraphics[width=1.0\linewidth]{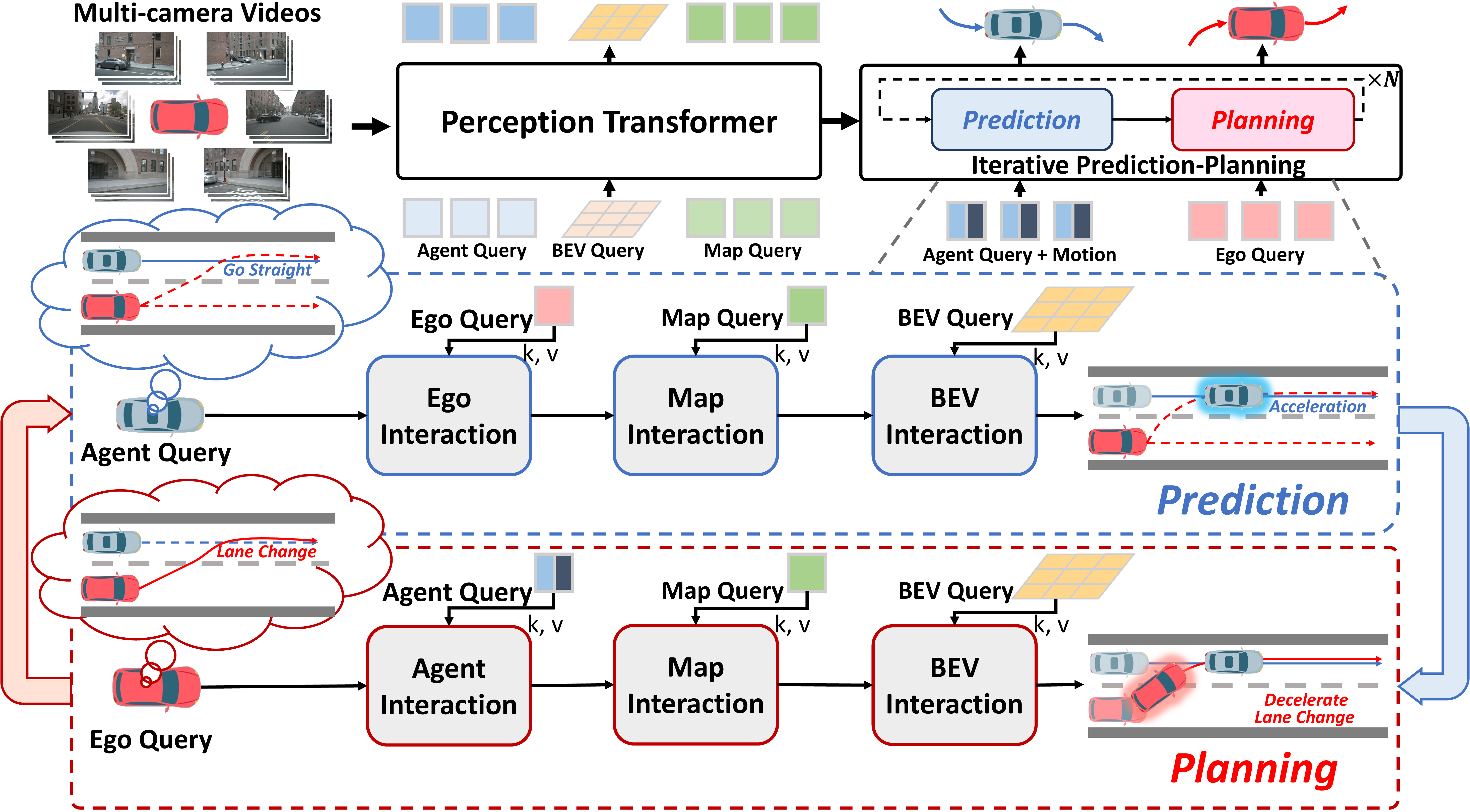}

\caption{\textbf{Overall architecture of our proposed self-driving framework, PPAD.} It consists of the Perception Transformer and the Iterative Prediction-Planning Module. The Perception Transformer encodes scene contexts into agent queries, map queries, and BEV queries. Then, the Prediction-Planning Module interleaves the processes of the agent motion prediction and the ego planning for $N$ times. Throughout the iterative Prediction and Planning processes, in-depth interactions are conducted among the ego, agents, map elements, and BEV features. In the Prediction process, the agent initially intends to go straight and is unaware of the potential motion of the ego. After interacting with the ego, map elements, and BEV features, the agent plans to be assertive and proceeds to accelerate. In the following Planning process, the ego knows the agent will accelerate through interacting with the updated agent query. It eventually plans to decelerate first and then conduct the lane change for safety reasons.}
\label{fig:overall}
\end{figure}

\section{Method}
\subsection{Framework Overview}
We present the overall framework, PPAD, in Fig.~\ref{fig:overall}, which comprises the principal modules of the Perception Transformer and our proposed Iterative Prediction-Planning Module. The Perception Transformer encodes the scene contexts into the BEV features map and further decodes as vectorized agents and map representations. The Iterative Prediction-Planning module consists of \textit{Prediction} and \textit{Planning} process in general. It dissects the dynamic interactions between the ego vehicle and the agents along the temporal dimension. Eventually, it predicts the motions of the agents and plans the future trajectory for the ego vehicle.

\noindent
\textbf{Image Features Module}
uses a shared image backbone network (e.g., ResNet \cite{he2016deep}) to extract image features for separate camera views.

\noindent
\textbf{BEV Features Module}
transform the semantic features from the multi-view cameras into a united bird's-eye-view. Specifically, we inherit the encoder from BEVFormer~\cite{li2022bevformer, yang2023bevformer} to construct the BEV features. The grid-shape learnable BEV queries $\mathbf{B}\in\mathbb{R}^{H\times W\times C}$ are randomly initialized and learned to interact with the multi-view image features through deformable attention~\cite{zhu2020deformable} to conduct spatial modeling. Temporal modeling is conducted in a recurrent manner, which applies the deformable attention between the current frame's BEV queries and the one from the previous time step.

\noindent
\textbf{Vectorized Features Module}
Inspired by the VAD~\cite{jiang2023vad} paradigm, we also encode the scene contexts into vectorized representations through a detection decoder head~\cite{li2022bevformer, zhu2020deformable} and a map element decoding head~\cite{liao2022maptr}, resulting in $N_A$ of learned agent queries $\mathbf{A}\in\mathbb{R}^{N_A\times C}$ and $N_M$ of learned map queries $\mathbf{M}\in\mathbb{R}^{N_M\times C}$. Separate MLP-based decoders will be attached to produce side output, which takes the learned queries as inputs and predicts with the agent attributes (locations, dimensions, classes, etc.) or map attributes (classes and map vectors described by points). Additionally, the agent queries will be combined with the learnable motion embeddings for modeling the diverse motions of the agents. The agents with motions are represented as $\mathbf{A}\in\mathbb{R}^{N_A\times N^{mot}_A\times C}$. Similarly, the ego vehicle is modeled with three modes, representing the high-level driving commands of {\em going straight, turn left, and turn right}, in the form of $\mathbf{E}\in\mathbb{R}^{N_E\times N^{mot}_E\times C}$.

\noindent
\textbf{Iterative Prediction-Planning Module} predicts the future trajectories for the ego vehicle and the agents in an interleaved fashion. Different from the traditional practice that predicts all the trajectories in one go, our PPAD framework articulates each step of motion planning by iterating the agent motion prediction and the ego planning processes. Thanks to the PPAD framework, we can conduct in-depth design to enforce key objects interactions (in Sec.~\ref{hierarchical}) on scene contexts in a coarse-to-fine manner. We further improve the driving performance for the ego vehicle by taking the noisy trajectory as each step prediction and training the PPAD framework to reconstruct its original position at the following time step (in Sec.~\ref{sec3.4}).

\subsection{Iterative Interactions of Prediction and Planning}
\label{sec3.2}
In the real world, the driving traffic changes constantly. Drivers plan and execute their decisions by ceaselessly reasoning the relationships among traffic participants in the scene. The planning task requires the self-driving system to have a good understanding of the scene and be capable of resolving the spatial-temporal causal factors. Therefore, we innovate the PPAD to dissect the planning task into multi-steps of the agent prediction and the ego planning processes and eventually promote consensuses among the ego's and the agents' future trajectories. The PPAD framework embodies the traffic interactions as gaming along space-time, producing a more accurate planning trajectory for the ego vehicle.

Specifically, the ego and agents inherit the same philosophy of alternatively optimizing their motion behavior based on each other's motion forecasting at each future time step. In the following section, we will demonstrate the agent prediction process and elaborate on the details of the ego planning process.

\subsubsection{Prediction Process}
As illustrated in Fig~\ref{fig:overall}, the agent predicts its subsequent step motion during the Prediction process, conditioned on the output of the ego vehicle's outcome from the previous Planning process. Specifically, the initial state of the agent query comprises its driving intention. It will then interact with the ego query updated from the previous planning process, which indicates the latest driving plan of the ego vehicle. After that, it will interact with map elements to choose the driving paths. At last, it gathers detailed geometric information by interacting with the BEV features and comes up with its precise next-step movement.

\subsubsection{Planning Process}
\label{sec3.2.2}
We consider the period of history with $T_{obs}$ steps and the future with $T_{fut}$ steps. The future trajectory of ego is denoted as $\{p^t_E\}^{t\in T_{fut}}$. For each agent ${a\in \mathbf{A}}$, the trajectory is represented as $\{p^{t}_a\}^{t\in T_{fut}}$. The positions of the detected map elements are denoted as $\{p_m\}_{m\in \mathbf{M}}$. We define the operator of $\mathcal{M}(p_E^t, p_{\mathbf{A}}/p_{\mathbf{M}}, s)$ to mask out the agents or map elements that are beyond a distance of $s$ towards the ego located at $p^t_E$ to conduct key objects attention, and we will discuss the algorithm details in Sec.~\ref{hierarchical}.

\noindent
\textbf{Agent Interaction}
The resulting agents' motions from the prediction process are located at $\{p^{t+1}_a\}_{a\in \mathbf{A}}$, comprising the motion states up to the time step of $t+1$. Moving the ego from $p^{t}_E$ to the future one-step of $p^{t+1}_E$, the ego vehicle should consider the agents' traffic globally and locally. From a more global perspective, the agents at a larger range provide more extensive information on traffic flow, which is essential in long-term trajectory planning. Regarding the spatial locality, the nearby agents are recognized as the key agents, which are supposed to be vitally related to the ego's driving decision. 

Therefore, we propose to conduct a hierarchical interaction with the agents through the attention mechanism to learn coarse-to-fine traffic context features for the ego. Centered at the ego's space-time position $p^{t}_E$, we initially define a distance set of $\mathbf{S}$ of $\{+\infty\ m, 15\ m, 7.5\ m\}$ which covers the coarse-to-fine perception ranges. We formed the multi-scale agent sets by applying $\mathcal{M}(p_E^t, p_{\mathbf{A}}, s)$ with different ranges. Then, the ego query interacts with the agents hierarchically through multi-head cross-attention, $\texttt{MHCA}$. We take the sum of the learned hierarchical attention results as the final values:
\begin{equation}
\label{eqn:1}
\mathbf{E}^{k} = \sum_{s\in \mathbf{S}}{\texttt{MHCA}(\mathbf{E}, \mathbf{A}^{k}, \mathcal{M}(p^t_\mathbf{E}, p^{t+1}_{\mathbf{A}^{k}},s))}, k\in[1,N_A^{mot}],
\end{equation}
where the ego independently queries information from different modes $k\in N_A^{mot}$ of agents, and then we stack the results output from different modes. Further, we apply the set operations to condense the features:
\begin{equation}
\label{eqn:2}
\mathbf{E}^\prime = \texttt{MAX}([\mathbf{E}^1,..,\mathbf{E}^{N_A^{mot}}]) + \texttt{MEAN}([\mathbf{E}^1,..,\mathbf{E}^{N_A^{mot}}]),
\end{equation}
where the \texttt{MAX} and \texttt{MEAN} are applied to aggregate features along the agents' mode dimension and output with the updated ego query $\mathbf{E}^\prime$.

\noindent
\textbf{Map Interaction}
Existing works~\cite{hu2023planning,jiang2023vad} tried to summarize all the required map information for planning by simply applying the global-level interaction once. They overlook the complexity of the evolving motion dynamic and overrate that the ego can plan precisely in the longer term by a single interaction with the map information.

With our proposed PPAD framework, we can enrich the ego-map interaction by considering the ego's local road conditions based on its latest position. This results in better identifying the useful map information for each step of planning. The ego query interacts with the map queries in a similar practice as the interaction with the agents. The difference is that the map instances are not movable in the future time steps. The local and global map information can be abstracted into the ego query by \texttt{MHCA}:
\begin{equation}
\label{eqn:3}
\mathbf{E}^{\prime\prime} = \sum_{s\in \mathbf{S}}{\texttt{MHCA}(\mathbf{E}^{\prime}, \mathbf{M}, \mathcal{M}(p^t_\mathbf{E}, p_{m},s))},
\end{equation}
where $\mathbf{E}^{\prime\prime}$ is the ego query updated with the critical map information for the next step of planning.

\noindent
\textbf{BEV Interaction}
BEV is the fundamental representation of the whole system, which is the abstraction of the multi-camera features. Beyond the vectorized representation, there is other non-structural environmental information, including roads and fences. UniAD~\cite{hu2023planning} models these non-structural pieces of stuff with an occupancy grid map. There are several drawbacks in UniAD: 1) occupancy grids consume large memory considering the whole scene's range. 2) UniAD failed to build the explicit interactions with the grid map. Therefore, we propose the BEV interactions that dynamically query the surrounding environment for each possible future step. This query process could help agents understand and learn the effects of their actions. Specifically, after applying the interactions above, the ego vehicle understands the dynamic agents' traffic better and knows its fronting road conditions. Nevertheless, planning a more precise motion requires the ego to comprehend the local detail geometric information. Hence, the ego query further interacts with BEV features to extract low-level geometric information. Specifically, we achieve by the deformable attention~\cite{zhu2020deformable}:
\begin{equation}
\label{eqn:4}
\mathbf{E}^{\prime\prime\prime} = \texttt{DeformAttn}(\mathbf{E}^{\prime\prime},p^t_\mathbf{E}, \mathbf{B}),
\end{equation}
where $p^t_\mathbf{E}$ is the location of ego at time $t$, and it serves as the reference point on the BEV features. The deformable attention $\texttt{DeformAttn}$ applies sparse attention around the reference points $p^t_\mathbf{E}$ and learns to pick up the low-level geometric information from the BEV for planning.

\noindent
\textbf{Motion Planning}
PPAD follows the same practice as ~\cite{hu2022st,hu2023planning,jiang2023vad}, which uses the information of the high-level driving commands: go straight, turn left, and turn right. The concatenated features of
$\mathbf{h_E}=[\mathbf{E}^{\prime},\mathbf{E}^{\prime\prime},\mathbf{E}^{\prime\prime\prime}]$ contain the information of the dynamic agent traffic, map semantics, and the precise environmental geometry. An \texttt{MLP} takes $\mathbf{h}_{\textbf{E}}$ as input and predicts the future one-step waypoint offset $w_\mathbf{E}^{t+1}=(x,y)$. We then update the ego state by applying another \texttt{MLP} on $\mathbf{h_E}$ for the next step of processing.

\begin{figure}[t!]
  \centering
  \begin{minipage}{.7\linewidth}
  
\begin{algorithm}[H]
\setstretch{1.2}
\caption{\small{Pseudo code of Key objects attention in a PyTorch-like style.}}
\label{alg:code}
\definecolor{codeblue}{rgb}{0.25,0.5,0.5}
\lstset{
	backgroundcolor=\color{white},
	basicstyle=\fontsize{8pt}{8pt}\ttfamily\selectfont,
	columns=fullflexible,
	breaklines=true,
	captionpos=b,
	commentstyle=\fontsize{8pt}{8pt}\color{codeblue},
	keywordstyle=\fontsize{8pt}{8pt}\color{codegreen},
	frame=tb,
}

\begin{lstlisting}[language=python]
################### initialization ###################
layer = nn.MultiheadAttention(embed_dim, num_head)
#################### forward pass ####################
def forward(layer, query, key, q_pos, k_pos, max_dis):
    # q_pos: position of the query with shape [B,Lq,2]
    # k_pos: position of the key with shape [B,Lk,2]
    # layer: attention layer as initialization
    # max_dis: distance threshold
    diff = (q_pos.unsqueeze(2) - k_pos.unsqueeze(1))
    dist = (diff ** 2).sum(-1).sqrt()
    attn_mask = (dist > max_dis).repeat(num_head, 1, 1)
    return attention(query, key, attn_mask=attn_mask)
    

\end{lstlisting}

\end{algorithm}

  \end{minipage}
\end{figure}

\subsection{Hierarchical Feature Learning}
\label{hierarchical}
Hierarchical structure has a better capability to capture and recognize fine-grained patterns. For the driving scenarios, the driving behavior is based on scene understanding both globally and locally. Driving tends to focus on only a few key objects, which demonstrates the spatial locality or local attention. Therefore, we design hierarchical \textbf{key objects attention} to exploit the coarse-to-fine scene contexts. Specifically, given a set of distance ranges, we first find the key objects (agents or map elements) within the given range. Consequently, we apply dynamic local attention, which only considers the interactions among agents or map elements in the local area. The pseudo code shown in Alg.~\ref{alg:code} delineates the implementation of dynamic key objects attention.

\subsection{Noisy Trajectory as Prediction}
\label{sec3.4}
PPAD interleaves the prediction and the planning processes to plan the ego and agents' trajectories step-by-step. Expert driving knowledge is then enforced into the model through imitation learning. Thanks to our multi-step framework and the inspiration from~\cite{li2022dn}, we introduce the noisy trajectory as the prediction to the PPAD while training. Specifically, we perturb each step of the ground truth ego trajectory by adding noise. The ego is then trained to predict the original next step waypoint offset of the ego regardless of disturbance on its starting noisy positions. The system is learned to predict the accurate waypoint offset by interacting with vectorized instances and the environment even though it starts at an inaccurate position. This strategy brings improvement to the planning performance.

\subsection{End-to-End Learning}
\noindent
\textbf{Scene Context Loss}
Similar to VAD~\cite{jiang2023vad}, we formulate the loss for the agents' motion and map as follows:
\begin{equation}
\label{eqn:5}
\mathcal{L}_{S}=\lambda_1\mathcal{L}_{agent}+\lambda_2\mathcal{L}_{map},
\end{equation}
where $\lambda_1$ and $\lambda_2$ are set as 1.0.

\noindent
\textbf{Constraint Loss}
Inspired by~\cite{jiang2023vad}, we propose the confidence-aware collision loss $\mathcal{L}_{CA-Col}$, which considers the potential collision of all the agents' motion modalities instead of only computing the loss on the agents' most confident mode. We multiply the resulting collision loss from each mode with the predicted confidence score. For the trajectories having the potential to collide, it will penalize more when the predicted confidence score is higher. Combined with ego-boundary overstepping $\mathcal{L}_{bd}$ and  ego-lane directional $\mathcal{L}_{dir}$ constraints proposed by~\cite{jiang2023vad}, the overall constraint loss is
\begin{equation}
\label{eqn:6}
\mathcal{L}_{C}=\lambda_3\mathcal{L}_{CA-Col}+\lambda_4\mathcal{L}_{bd}+\lambda_5\mathcal{L}_{dir},
\end{equation}
where $\lambda_3$ and $\lambda_4$ are set as 1.0, and $\lambda_5$ is set as 0.5.

\noindent
\textbf{Planning Loss}
We conduct $L_1$ loss between each step of the ego's prediction ${w}_\mathbf{E}^{t}$ and the ground truth's waypoint offset $\Tilde{w}_\mathbf{E}^{t}$ along the future time horizon:
\begin{equation}
\label{eqn:7}
\mathcal{L}_{Plan} = \frac{1}{T_{fut}}\sum^{T_{fut}}_{t=1}{\lVert w_\mathbf{E}^{t}-\Tilde{w}_\mathbf{E}^{t}\rVert}_1.
\end{equation}

The overall end-to-end trainable loss function is formed by the sum of the perception loss, constraint loss, and planning loss. The same constraint losses $\mathcal{L}_{C}^{noisy}$ and planning losses $\mathcal{L}_{plan}^{noisy}$ will be applied to the predictions taking the noisy trajectories as input:
\begin{equation}
\label{eqn:8}
\mathcal{L} = \mathcal{L}_{S} + \zeta_1(\mathcal{L}_{C} + \mathcal{L}_{Plan}) +\zeta_2(\mathcal{L}_{C}^{noisy} + \mathcal{L}_{Plan}^{noisy}),
\end{equation}
where $\zeta_1$ is set as 0.6 and $\zeta_2$ is set as 0.4.

\begin{table}[t!]
\begin{center}
\resizebox{1.0\textwidth}{!}{
\setlength{\tabcolsep}{0.3em}
\renewcommand{\arraystretch}{1.05}
\begin{tabular}{c|l|cccc|cccc|cc}
\Xhline{2.0\arrayrulewidth}

\multirow{2}{*}{} &
\multirow{2}{*}{Method} &
\multicolumn{4}{c|}{L2 (m) $\downarrow$} & 
\multicolumn{4}{c|}{Collision (\%) $\downarrow$} &
\multirow{2}{*}{Latency (ms)} &
\cellcolor{gray!30} \\
& & 1s & 2s & 3s & \cellcolor{gray!30}Avg. & 1s & 2s & 3s & \cellcolor{gray!30}Avg. & & \cellcolor{gray!30}\multirow{-2}*{FPS} \\
\Xhline{2.0\arrayrulewidth}

\multirow{7}{*}{\textbf{ST-P3 Metrics}} 

& ST-P3~\cite{hu2022st} 
& 1.33 & 2.11 & 2.90 & \cellcolor{gray!30}2.11 & 0.23 & 0.62 & 1.27 & \cellcolor{gray!30}0.71 & 628.3 & \cellcolor{gray!30}1.6 \\

& VAD-Tiny~\cite{jiang2023vad} 
& 0.46 & 0.76 & 1.12 & \cellcolor{gray!30}0.78 & 0.21 & 0.35 & 0.58 & \cellcolor{gray!30}0.38 & \textbf{59.5} & \cellcolor{gray!30}\textbf{16.8} \\

& VAD-Base~\cite{jiang2023vad} 
& {0.41} & {0.70} & {1.05} & \cellcolor{gray!30}{0.72} & \textbf{0.07} & {0.17} & {0.41} & \cellcolor{gray!30}{0.22} & 224.3 & \cellcolor{gray!30}4.5 \\

& OccNet~\cite{sima2023scene} 
& 1.29& 2.13& 2.99 & \cellcolor{gray!30}{2.13} & 0.21 & 0.59 & 1.37 & \cellcolor{gray!30}{0.72} & - & \cellcolor{gray!30}- \\

& FusionAD~\cite{ye2023fusionad} 
& {-} & {-} & {-} & \cellcolor{gray!30}{1.03} & 0.25 & 0.13 & \textbf{0.25} & \cellcolor{gray!30}{0.21} & - & \cellcolor{gray!30}- \\

% \Xhline{2.0\arrayrulewidth}

\rowcolor{gray!15} & \textbf{Ours (Progress.)} & 0.31 & 0.56 & 0.87 & \cellcolor{gray!30}{0.58} & 0.08 & \textbf{0.12} & 0.38 & \cellcolor{gray!30}{\textbf{0.19}} & {385} & \cellcolor{gray!30}{2.6}  \\

\rowcolor{gray!15} & \textbf{Ours} & \textbf{0.25} & \textbf{0.45} & \textbf{0.73} & \cellcolor{gray!30}\textbf{0.48} & \textbf{0.07} & {0.15} & 0.36 & \cellcolor{gray!30}\textbf{0.19} & {385} & \cellcolor{gray!30}{2.6}  \\

\Xhline{2.0\arrayrulewidth}

\multirow{8}{*}{\textbf{UniAD Metrics}} 

& NMP$^\dagger$~\cite{zeng2019nmp} 
& - & - & 2.31 & \cellcolor{gray!30}- & - & - & 1.92 & \cellcolor{gray!30}- & - & \cellcolor{gray!30}- \\

& SA-NMP$^\dagger$~\cite{zeng2019nmp} 
& - & - & 2.05 & \cellcolor{gray!30}- & - & - & 1.59 & \cellcolor{gray!30}- & - & \cellcolor{gray!30}- \\

& FF$^\dagger$~\cite{hu2021safe}
& 0.55 & 1.20 & 2.54 & \cellcolor{gray!30}1.43 & 0.06 & 0.17 & 1.07 & \cellcolor{gray!30}0.43 & - & \cellcolor{gray!30}- \\

& EO$^\dagger$~\cite{khurana2022differentiable} 
& 0.67 & 1.36 & 2.78 & \cellcolor{gray!30}1.60 & 0.04 & \textbf{0.09} & 0.88 & \cellcolor{gray!30}0.33 & - & \cellcolor{gray!30}- \\

& UniAD~\cite{hu2023planning} 
& 0.48 & 0.96 & 1.65 & \cellcolor{gray!30}1.03 & 0.05 & {0.17} & \textbf{0.71} & \cellcolor{gray!30}\textbf{0.31} & 555.6 & \cellcolor{gray!30}1.8 \\

& VAD-Base~\cite{jiang2023vad} 
& 0.50 & 1.02 & 1.69 & \cellcolor{gray!30}{1.07} & \textbf{0.00} & {0.30} & {0.95} & \cellcolor{gray!30}{0.42} & 224.3 & \cellcolor{gray!30}4.5 \\

% \Xhline{2.0\arrayrulewidth}
\rowcolor{gray!15} & \textbf{Ours (Progress.)} & 0.38 & 0.83 & 1.45 & \cellcolor{gray!30}{0.89} & 0.02 & 0.20 & 0.93 & \cellcolor{gray!30}{0.38} & {385} & \cellcolor{gray!30}{2.6}  \\

\rowcolor{gray!15} & \textbf{Ours} & \textbf{0.30} & \textbf{0.69} & \textbf{1.26} & \cellcolor{gray!30}\textbf{0.75} & 0.03 & {0.22} & 0.73 & \cellcolor{gray!30}{0.33} & {385} & \cellcolor{gray!30}{2.6}  \\

\Xhline{2.0\arrayrulewidth}
\end{tabular} }
\end{center}

\caption{
\textbf{Open-loop planning results on the nuScenes dataset~\cite{caesar2020nuscenes}.} The results of other methods are obtained from the original paper. We faithfully re-evaluate VAD~\cite{jiang2023vad} based on the UniAD~\cite{hu2023planning} metrics. As for our PPAD, we provide two versions of results that utilize different training strategies. Progress. means that we follow the progressive training pipeline as proposed in VAD~\cite{jiang2023vad}, which trained all tasks except the planning task in the first 48 epochs and then finetuned with another 12 epochs for the planning task. The second row for our method trains the whole network for 60 epochs and then finetuned for another 12 epochs incorporating noisy trajectories. The latency of ST-P3~\cite{hu2022st}, VAD~\cite{jiang2023vad}, and ours are measured on one NVIDIA Geforce RTX 3090 GPU, while UniAD is measured on one NVIDIA Tesla A100 GPU.
}
\label{tab:sota-plan}

\end{table}

\begin{table}[t!]
\begin{center}
\resizebox{0.65\textwidth}{!}{
    \setlength{\tabcolsep}{0.3em}
      \renewcommand{\arraystretch}{1.05}
\begin{tabular}{l|cccc|cccc}
\Xhline{2.0\arrayrulewidth}

\multirow{2}{*}{Method} &
\multicolumn{4}{c|}{L2 (m) $\downarrow$} & 
\multicolumn{4}{c}{Collision (\%) $\downarrow$} \\

& 1s & 2s & 3s & \cellcolor{gray!30}Avg. & 1s & 2s & 3s & \cellcolor{gray!30}Avg. \\
\Xhline{2.0\arrayrulewidth}

VAD~\cite{jiang2023vad} 
& {1.04} & {1.79} & {2.60} & \cellcolor{gray!30}{1.81} & {0.61} & {1.26} & {2.25} & \cellcolor{gray!30}{1.37} \\

\rowcolor{gray!15} \textbf{Ours} 
& \textbf{0.73} & \textbf{1.30} & \textbf{1.98} & \cellcolor{gray!30}\textbf{1.34} & \textbf{0.51} & \textbf{0.85} & \textbf{1.31} & \cellcolor{gray!30}\textbf{0.89} \\

\Xhline{2.0\arrayrulewidth}
\end{tabular} 
}
\end{center}

\caption{
\textbf{Open-loop planning results on the Argoverse dataset~\cite{caesar2020nuscenes}.} The results are evaluated under the ST-P3~\cite{hu2022st} and VAD~\cite{jiang2023vad} metrics.
}
\label{tab:argo-plan-test}

\end{table}

\begin{table}[t!]

\begin{center}
\centering
\resizebox{0.8\textwidth}{!}{
\setlength{\tabcolsep}{0.3em}
\renewcommand{\arraystretch}{1.05}
\begin{tabular}{l|cc|c|ccc}

\Xhline{2.0\arrayrulewidth}
\multirow{2}{*}{Method} & \multicolumn{2}{c|}{Detection} & {Map} & \multicolumn{3}{c}{Motion Forecasting}\\
& NDS $\uparrow$& mAP$\uparrow$ & mAP$\uparrow$ & minADE (m)$\downarrow$ & minFDE (m) $\downarrow$ & MR$\downarrow$ \\

\Xhline{2.0\arrayrulewidth}
VAD~\cite{jiang2023vad} &0.459 & 0.329& 0.476 & 0.678& \textbf{0.882}& 0.08 
\\
UniAD~\cite{hu2023planning} &\textbf{0.499} & \textbf{0.382} & - & 0.708& 1.02 &0.13 \\
\Xhline{2.0\arrayrulewidth}
\textbf{Ours} &0.465 & 0.332 & \textbf{0.519} & \textbf{0.676}& 0.889 & \textbf{0.07} \\
\Xhline{2.0\arrayrulewidth}

\end{tabular}
}
\end{center}

\caption{
 Results comparison on the tasks beyond the planning task.
}
\label{tab:all_metrics}

\end{table}

\section{Experiments}
\subsection{Experimental Setup}
\noindent\textbf{Dataset} 
We evaluate our method on two challenging large-scale real-world datasets, nuScenes~\cite{caesar2020nuscenes} and Argoverse2~\cite{wilson2023argoverse}. We conduct ablation studies to evaluate the effectiveness of our proposed components on the nuScenes~\cite{caesar2020nuscenes} dataset. The nuScenes dataset~\cite{caesar2020nuscenes} provides about $1$K $20$-second diverse driving scenes collected in Boston, Pittsburgh, Las Vegas, and Singapore for open-loop settings. Key samples which contain $6$ camera images are annotated at $2$Hz. The Argoverse2 dataset comprises $1$K $15$-second scenes. It captures $7$ RGB images on each frame. We align the data sampling frequency to nuScenes~\cite{caesar2020nuscenes} by downsampling the annotated frames by an interval of $5$ frames.

\noindent\textbf{Metrics} We adopt the metrics of L2 Displacement Error (L2) and Collision Rate (CR) for evaluation~\cite{hu2022st}. L2 is measured between the prediction and ground-truth trajectories over the timesteps $1$-s, $2$-s, and $3$-s, evaluating the trajectory quality. Collision Rate (CR) measures how often the collision occurs between the ego vehicle and the other agents along the planning horizon, reflecting the trajectory safety. We notice that the UniAD~\cite{hu2023planning} adheres to different calculations from ST-P3~\cite{hu2022st} and VAD~\cite{jiang2023vad}: the former~\cite{hu2023planning} reports the evaluations at each second. In contrast, the latters~\cite{hu2022st,jiang2023vad} reports the results of the cumulative average by each second. We faithfully make comparisons of our method to others with these two kinds of computations.

\noindent\textbf{Implementation Details}
We strictly follow the standard settings as proposed by UniAD~\cite{hu2023planning} and VAD~\cite{jiang2023vad}, which did not use the information of the historical ego trajectory. Same as VAD~\cite{jiang2023vad}, the perception range is $60m \times 30m$. PPAD also recurrently encodes $2$-s historical information into BEV and predicts the $3$-s trajectory in the future. It also conducts the tasks of motion prediction and map construction. Our PPAD trains with a batch size of $1$ using the Adam~\cite{kingma2014adam} optimizer set with an initial learning rate of 2e-4. It takes us about six days to conduct the end-to-end training on eight NVIDIA A30 24GB GPUs.

\subsection{Main Results}
\noindent
\textbf{Planning Results} As shown in Tab.~\ref{tab:sota-plan}, our PPAD outperforms the current state-of-the-art performance by a large margin. Especially for the L2 distance metrics, there are about $20$\% of consistent improvements can be observed along the temporal horizon. Thanks to the iterative interaction of prediction and planning, PPAD can help avoid collisions, leading to better results on collision rate compared to the one-shot representative VAD~\cite{jiang2023vad}. At the same time, we maintain a competitive efficiency compared to UniAD~\cite{hu2023planning}.

We further make a fair comparison between our method and the baseline, VAD~\cite{jiang2023vad}, on the other dataset of Argoverse 2~\cite{chang2019argoverse,wilson2023argoverse}. In Tab.~\ref{tab:argo-plan-test}, our method can consistently outperform the baseline with a clear margin in both L2 distance and collision rate metrics.

\noindent
\textbf{Subtasks Results} To demonstrate the overall performance of our PPAD, we also provide the evaluation results besides planning metrics on the traditional perception and motion forecasting task in Tab.~\ref{tab:all_metrics}. Our PPAD also achieves promising performance in upstreaming perception and prediction tasks, which demonstrates that the whole system is jointly optimized.

\begin{figure}[t!]
\centering
\includegraphics[width=0.9\linewidth]{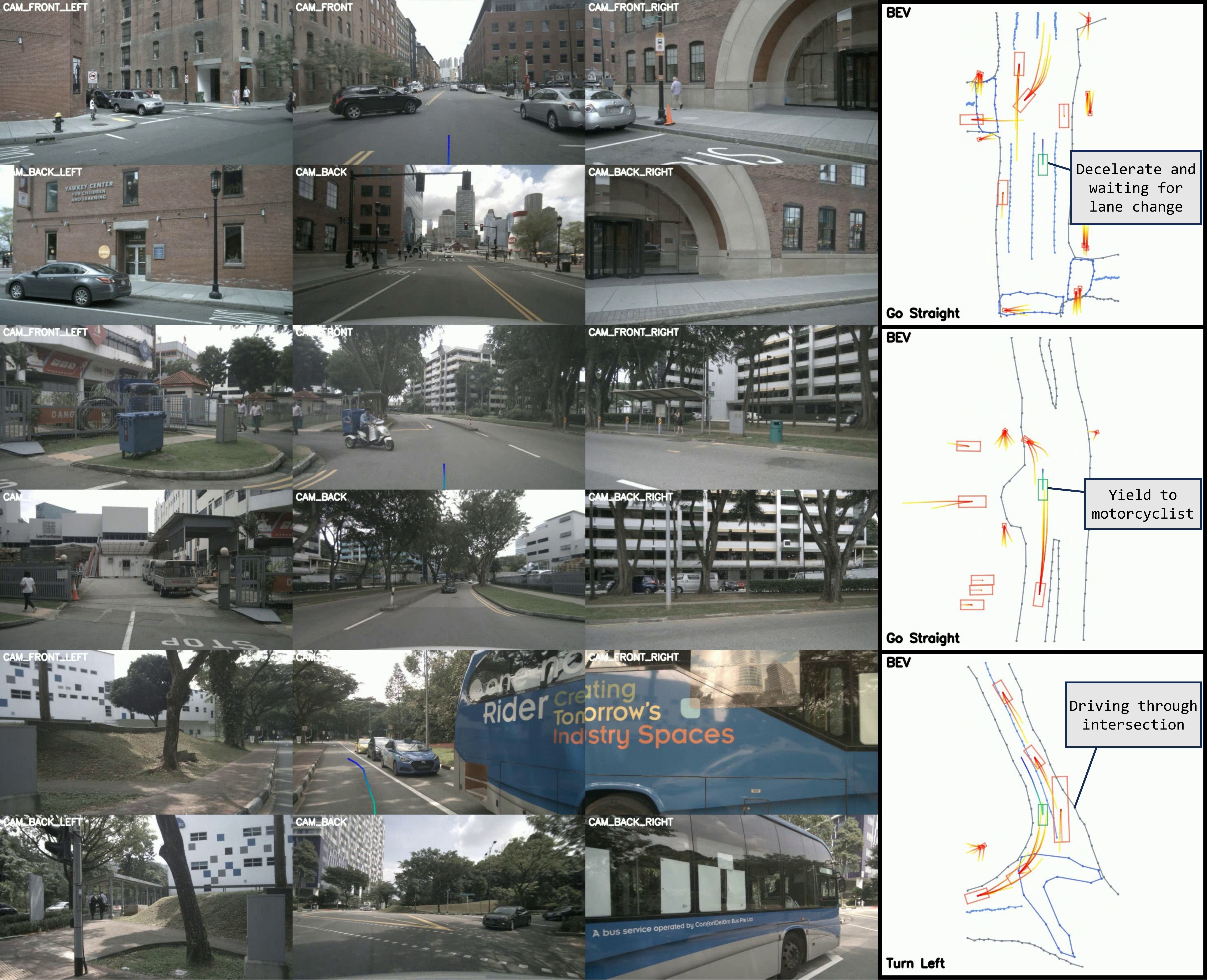}

\caption{\textbf{Qualitative results of PPAD.} The green box in the figure demonstrates the ego agent, while the red ones are agents.}
\label{fig:qualitive_results}
\end{figure}

\noindent
\textbf{Qualitative Results} We provide qualitative results shown in Fig.~\ref{fig:qualitive_results}. PPAD can perceive the scene precisely and predict with reasonable and diverse motions for the surrounding agents. It also plans a smooth and accurate trajectory for the ego vehicle.

\subsection{Ablation Study}
\label{sec4.3}
The following experiments adhere to the progressive training pipeline as proposed in VAD~\cite{jiang2023vad}.

\noindent
\textbf{Effectiveness of Designs}
We provide ablation studies to verify the effectiveness of our proposed components. As shown in Tab.~\ref{tab:design}, the proposed PPAD framework (row 2) brings a remarkable improvement compared with one-shoot methods~\cite{jiang2023vad} (row 1). The multi-step interactions help the ego agent better understand the intention and potential effects brought by its actions along the temporal horizon, leading to an over $10$\% L2 distance error reduction. We can observe a further improvement regarding the L2 distance with our proposed key objects attention (row 3). The slight degradation in collision rate might be due to the key objects attention being conducted on the ego with each mode of the agents, and the diverse modes of the agents mislead the behavior of the ego vehicle. From rows 5-7, the confidence-aware collision loss and noisy trajectory as the prediction can circumvent this phenomenon and further escalate the capability in planning accuracy and avoiding a collision. When we keep all of the remaining components while not applying key objects attention (row 4), we observed the degradations from row 7 in L2 (0.58 m vs. 0.59 m) and collision rate (0.19\% vs. 0.21\%), proving the effectiveness of the key objects attention.

\begin{table}[t!]

\begin{center}
\centering
\resizebox{\textwidth}{!}{
\setlength{\tabcolsep}{0.3em}
% \renewcommand{\arraystretch}{1.5}
% \resizebox{0.9\textwidth}{!}{
      \renewcommand{\arraystretch}{1.05}
\begin{tabular}{c|cc|cc|cccc|cccc|c}

\Xhline{2.0\arrayrulewidth}
\multirow{2}{*}{} & \multirow{2}{*}{PPAD} & \multirow{2}{*}{Key Objects Attn.} & \multirow{2}{*}{$\mathcal{L}_{CA-Col}$} & \multirow{2}{*}{Noisy Traj.} & \multicolumn{4}{c|}{L2 (m) $\downarrow$} &\multicolumn{4}{c|}{Collision (\%) $\downarrow$}  & \multirow{2}{*}{Latency (ms)} \\
&  &  &  &   & 1s & 2s & 3s & \cellcolor{gray!30}Avg. & 1s & 2s & 3s & \cellcolor{gray!30}Avg. \\

\Xhline{2.0\arrayrulewidth}
 1 & - & - & - & - 
 & {0.41} & {0.70} & {1.05} & \cellcolor{gray!30}{0.72} & \textbf{0.07} & {0.17} & {0.41} & \cellcolor{gray!30}{0.22}
 & 224 \\ 
 
 2 & \cmark & - & - & - 
 & 0.35 & 0.59 & 0.88 & \cellcolor{gray!30}{0.60} & 0.11 & {0.18} & {0.41} & \cellcolor{gray!30}{0.23}
 & 318 \\
 
 3 & \cmark & \cmark & - & - 
 & 0.32 & \textbf{0.55} & \textbf{0.83} & \cellcolor{gray!30}{\textbf{0.57}} & 0.19 & {0.28} & {0.58} & \cellcolor{gray!30}{0.35} 
& 385 \\

 4 & \cmark & - & \cmark & \cmark
 & 0.33 & 0.57 & 0.88 & \cellcolor{gray!30}{0.59} & 0.10 & {0.14} & {0.38} & \cellcolor{gray!30}{0.21}
 & 318 \\
 
\cline{1-14}
 5 & \cmark & \cmark & \cmark & - 
 & 0.35	& 0.59 &0.89 & \cellcolor{gray!30}{0.61} & 0.08 & {0.14} & \textbf{0.32} & \cellcolor{gray!30}{\textbf{0.18}}
 & 385 \\
 
 6 & \cmark & \cmark & - & \cmark 
 & 0.34 & 0.59 & 0.90 & \cellcolor{gray!30}{0.61} & 0.10 & {0.14} & {0.34} & \cellcolor{gray!30}{0.19}
 & 385 \\

 7 & \cmark & \cmark & \cmark & \cmark 
 & \textbf{0.31} & 0.56 & 0.87 & \cellcolor{gray!30}{0.58} & 0.08 & \textbf{0.12} & {0.38} & \cellcolor{gray!30}{0.19}
 & 385 \\

\Xhline{2.0\arrayrulewidth}
\end{tabular}
}
% }
\end{center}

\caption{
\textbf{Component study for PPAD.} Models follow the progressive training pipeline. PPAD means the auto-regressive framework with the designed stepwise interactions. Key Objects Attention represents hierarchical feature learning for the key objects. $\mathcal{L}_{CA-Col}$ represents the loss design for the confidence-aware collision loss. Noisy Traj. means that we incorporate noisy trajectories while training.
}
\label{tab:design}

\end{table}

\noindent
\textbf{Effectiveness of Interactions}
Our PPAD framework enables richer interactions among scene contexts, introducing local and global understandings of the world to the model. We further conduct ablation studies to demonstrate the performance gains brought by the interactions. We conduct the ablation study on interactions under the setting (Tab.~\ref{tab:design} in row 5) without using the noisy trajectory as the prediction for better comparison. As illustrated in Tab.~\ref{tab:interaction}, we can achieve the best performance in terms of L2 distance and collision by incorporating all of the interactions.

\begin{table}[t!]
\begin{center}
\centering
\setlength{\tabcolsep}{0.3em}
% \resizebox{0.475\textwidth}{!}{
\resizebox{0.6\textwidth}{!}{
\renewcommand{\arraystretch}{1.05}
\begin{tabular}{ccc|cccc|cccc}
    \Xhline{2.0\arrayrulewidth}
    \multirow{2}{*}{EA} & \multirow{2}{*}{Map} & \multirow{2}{*}{BEV} & \multicolumn{4}{c|}{L2 (m) $\downarrow$} &\multicolumn{4}{c}{Collision (\%) $\downarrow$}  \\
    &  &  & 1s & 2s & 3s & \cellcolor{gray!30}Avg. & 1s & 2s & 3s & \cellcolor{gray!30}Avg. \\
    \Xhline{2.0\arrayrulewidth}
    
    \cmark & - & - &
    0.35 & 0.58 & 0.89 & \cellcolor{gray!30}{0.61} & 0.23 & 0.24 & 0.49 & \cellcolor{gray!30}{0.32} \\ 
    
    \cmark & \cmark & - &
    0.38 & 0.64 & 0.95 & \cellcolor{gray!30}{0.66} & 0.17 & 0.18 & 0.48 & \cellcolor{gray!30}{0.28} \\ 
    
    \Xhline{2.0\arrayrulewidth}
    \cmark & \cmark & \cmark 
    & 0.35	& 0.59 &0.89 & \cellcolor{gray!30}{0.61} & 0.08 & 0.14 & 0.32 & \cellcolor{gray!30}{0.18} \\

    \Xhline{2.0\arrayrulewidth}
\end{tabular}
}
\end{center}

\caption{
\textbf{Interaction study for PPAD.} EA, Map, BEV mean the interactions of the ego with agents, the ego and agents with the map, the ego and agents with the BEV, respectively.
}
\label{tab:interaction}
\end{table}

\noindent
\textbf{Effect of Different Iterations on Prediction and Planning Interaction}
Our innovative PPAD interaction mechanism not only better models motion planning as gaming among the ego and the agents but also enriches the ego's / agents' interactions with their local environments. We further conducted the ablation study on the different iterations in applying the interactions of prediction and planning, as shown in Tab.~\ref{tab:steps}. Specifically, PPAD plans the trajectories with $3$, $2$, and $1$ steps of waypoints after each of the prediction-planning processes for the interaction iterations of $2$, $3$, and $6$ in Tab.~\ref{tab:steps}. It is demonstrated that the performance reaches the best as we conduct the interactions of prediction and planning processes at every future step.

\begin{table}[t!]

\begin{center}
\centering
\resizebox{\textwidth}{!}{
% \resizebox{0.9\textwidth}{!}{
\setlength{\tabcolsep}{0.3em}
      \renewcommand{\arraystretch}{1.05}
\begin{tabular}{c|cccc|cccc|c}

\Xhline{2.0\arrayrulewidth}
\multirow{2}{*}{Interaction Iterations} & \multicolumn{4}{c|}{L2 (m) $\downarrow$} &\multicolumn{4}{c|}{Collision (\%) $\downarrow$} & \multirow{2}{*}{Latency (ms)} \\
& 1s & 2s & 3s & \cellcolor{gray!30}Avg. & 1s & 2s & 3s & \cellcolor{gray!30}Avg.\\
\Xhline{2.0\arrayrulewidth}

\multicolumn{1}{l|}{2\quad(Every 1.5 sec)} &
0.36 & 0.63 & 0.96 & \cellcolor{gray!30}{0.65} & 0.14 & 0.20 & 0.43 & \cellcolor{gray!30}{0.25} & 306 \\ 

\multicolumn{1}{l|}{3\quad(Every 1.0 sec)} &
0.37 & 0.63 & 0.95 & \cellcolor{gray!30}{0.65} & 0.10 & 0.15 & 0.39 & \cellcolor{gray!30}{0.21} & 326 \\

\Xhline{2.0\arrayrulewidth}
\multicolumn{1}{l|}{6\quad(Every 0.5 sec)} &
\textbf{0.31} & \textbf{0.56} & \textbf{0.87} & \cellcolor{gray!30}{\textbf{0.58}} & \textbf{0.08} & \textbf{0.12} & \textbf{0.38} & \cellcolor{gray!30}{\textbf{0.19}} & 385 \\

\Xhline{2.0\arrayrulewidth}
\end{tabular}
}
% }
\end{center}

\caption{
The ablation study of conducting different interaction iterations in the future time horizons.
}
\label{tab:steps}

\end{table}

\section{Conclusion}
In this paper, we have presented a novel autonomous driving framework, \textbf{PPAD}. Different from the previous methods that lack in-depth modeling of interactions, we pose the planning problem as a multi-step \textbf{Prediction} and \textbf{Planning} gaming process among the ego vehicle and agents. With \textbf{PPAD} architecture, our proposed hierarchical dynamic key objects attention is incorporated to learn local and global scene contexts at each step and eventually plan with a more precise trajectory. The confidence-aware collision constraint and noisy trajectories are utilized while training to improve driving safety further. In general, our proposed novel \textbf{PPAD} achieves compelling performance upon the existing state-of-the-art methods, and we hope the \textbf{PPAD} framework can inspire the community to further exploration.

\clearpage  % TODO REVIEW/FINAL: This \clearpage needs to be removed from both review and camera-ready versions.

\section*{Acknowledgements}
The authors are thankful for the financial support from the Hetao Shenzhen-HongKong Science and Technology Innovation Cooperation Zone (HZQB-KCZYZ-2021055), this work was also supported by Shenzhen Deeproute.ai Co., Ltd (HZQB-KCZYZ-2021055).

\bibliographystyle{splncs04}
\bibliography{main}
\end{document}